\documentclass{article}

\PassOptionsToPackage{numbers, compress}{natbib}

\usepackage[nonatbib, preprint]{nips_2018}

\usepackage[utf8]{inputenc} 
\usepackage[T1]{fontenc}    
\usepackage{hyperref}       
\usepackage{url}            
\usepackage{booktabs}       
\usepackage{amsfonts}       
\usepackage{nicefrac}       
\usepackage{microtype}      

\usepackage{graphicx}
\usepackage{subfigure}
\usepackage{adjustbox}
\usepackage[symbol]{footmisc}
\usepackage{changepage}

\title{How Secure are Deep Learning Algorithms from Side-Channel based Reverse Engineering?}

\author{
  Manaar Alam \\
  Indian Institute of Technology Kharagpur\\
  Kharagpur, West Bengal, India. \\
  \texttt{alam.manaar@iitkgp.ac.in} \\
  \And
  Debdeep Mukhopadhyay \\
  Indian Institute of Technology Kharagpur\\
  Kharagpur, West Bengal, India. \\
  \texttt{debdeep@iitkgp.ac.in}
}

\begin{document}

\maketitle

\begin{abstract}
Deep Learning algorithms have recently become the de-facto paradigm for various prediction problems, which include many privacy-preserving applications like online medical image analysis. Presumably, the privacy of data in a deep learning system is a serious concern. There have been several efforts to analyze and exploit the information leakages from deep learning architectures to compromise data privacy. In this paper, however, we attempt to provide an evaluation strategy for such information leakages through deep neural network architectures by considering a case study on Convolutional Neural Network (CNN) based image classifier. The approach takes the aid of low-level hardware information, provided by Hardware Performance Counters (HPCs), during the execution of a CNN classifier and a simple hypothesis testing in order to produce an alarm if there exists any information leakage on the actual input.
\end{abstract}

\section{Introduction}
Over the past few years, we have seen an outburst of deep learning research in both industry and academia, as it considerably outperforms prior machine learning based techniques in a wide variety of domains, such as image recognition~\cite{russakovsky2015imagenet}, machine translation~\cite{hermann2014multilingual, bahdanau2014neural}, speech processing~\cite{graves2013speech}, etc. Convolutional Neural Networks (CNN) have been widely used successfully for image recognition tasks, starting from identifying plant and animal species~\cite{chen2014deep} to autonomous driving~\cite{chen2015deepdriving}. CNN is also recently being used for many other privacy-preserving applications like online medical image analysis~\cite{tajbakhsh2016convolutional, anwar2018medical}, in which the data privacy of any user requires utmost attention. There have been some recent attempts of reverse engineering and retrieving the parameters of Neural Network by side-channel information~\cite{hua2018reverse,yan2018cache,batina2018csi} in order to steal the model and jeopardize the privacy. Wei \emph{et al.}~\cite{wei2018know} presented an approach which goes one step further even to determine the input image of an FPGA-based CNN implementation by observing power side-channel. This growing efforts to reverse engineer the CNN inputs would directly hamper the privacy of users, which could lead to severe issues, like discrimination, safety violations, etc. Moreover, it could lead to substantial economic losses for companies if such private data which is used to train the CNNs is revealed to other parties, which could include competitors. Hence, in this paper, we pursue an effort to develop an evaluation strategy for such divulging of neural network inputs. The target scenario in our approach is when a CNN is executing on a standard desktop, and the \emph{evaluator} is expected to perform a dynamic evaluation and throw alarms when it detects possibilities of such leakages when the CNN classifies its inputs. The tool that the \emph{evaluator} employs for this detection is a set of registers which are called Hardware Performance Counters (HPCs), present in most of the computing platforms, ranging from workstations to embedded processors. The data acquired from the HPCs are run-time monitored by the \emph{evaluator}, which computes $t$-statistics of the data to notify that the ongoing CNN classifier is emanating side channel information which is significant for adversaries to be able to determine the input even treating the CNN implementation as a black-box.

\subsection*{Contribution}
We strive to provide an evaluation strategy in order to measure private information leakages, i.e., the actual input to the deep neural network architecture during its prediction operation using readily available Hardware Performance Counters and basic hypothesis testing methodology, which to the best of our knowledge has not been attempted so far.

\section{Motivation behind the Work}
The process of classification by any DNN based classifier consists of a series of multiplication and addition operations that it executes on the computing environment (i.e., CPUs or GPUs)\footnote[1]{In this work, we focus on the CPU implementation of an image classifier based on CNN.}. It is a well established fact in the literature that the execution of any process on the CPU leaks valuable side-channel information through processor cache, branch predictor unit and other low-level hardware activities~\cite{ge2016your}. The motivation behind this work is to explore the possibility of private information leakages in terms of these hardware events during the classification operation of a CNN before deploying it in a large-scale application.

\begin{figure}[!t]
    \centering
	\subfigure[]{
		\includegraphics[width=0.4\textwidth]{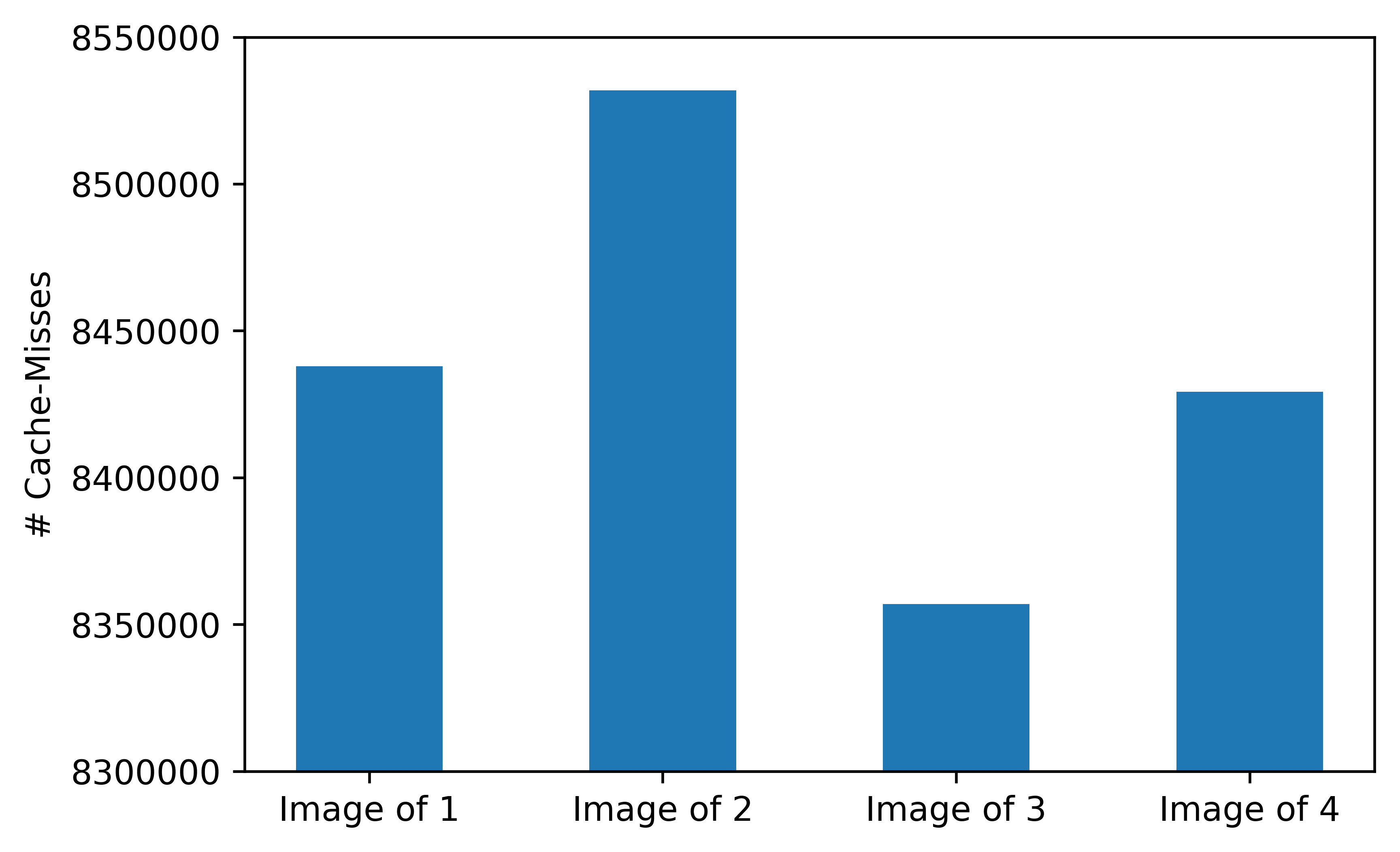}
	}
	\subfigure[]{
		\includegraphics[width=0.4\textwidth]{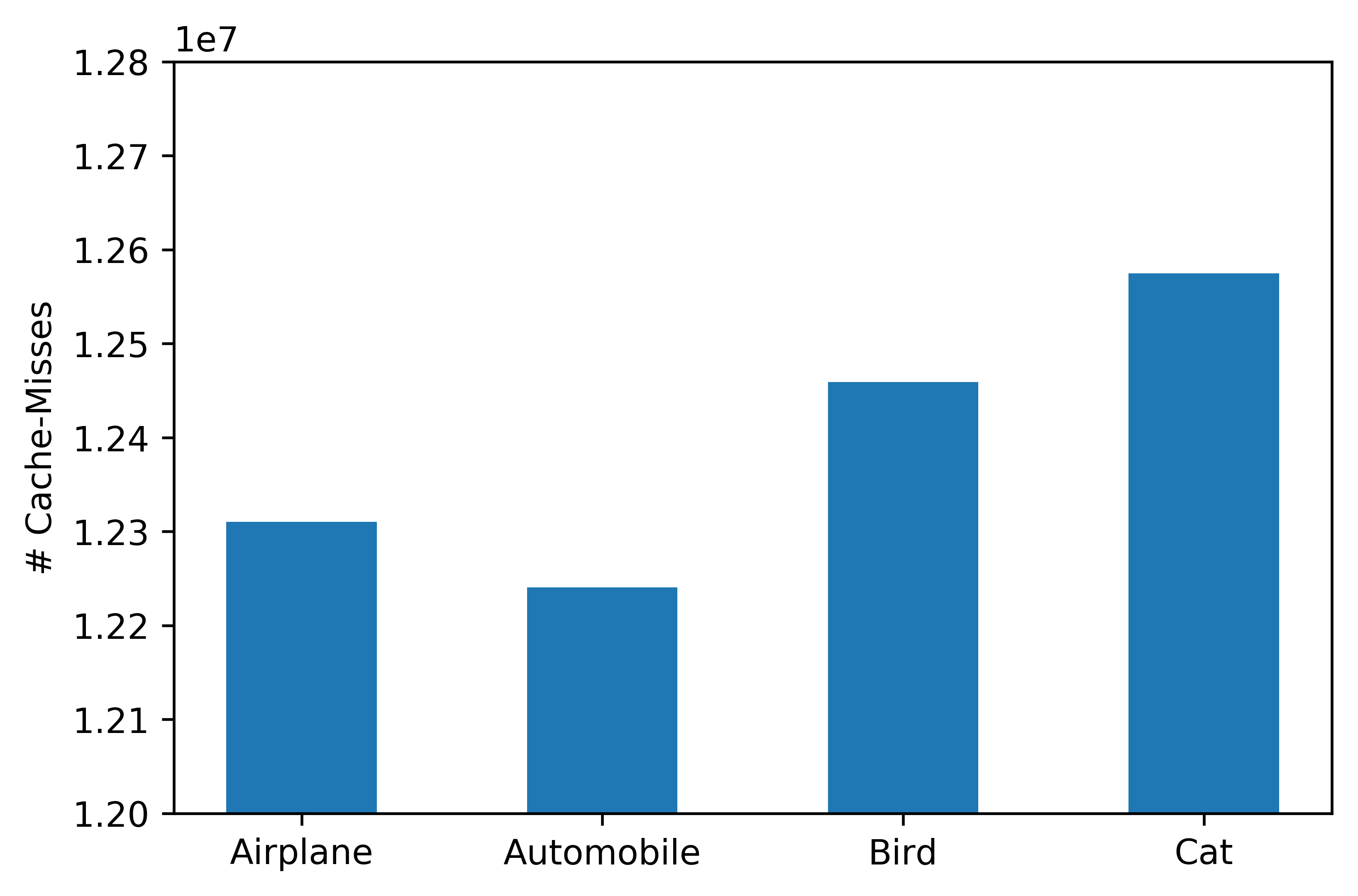}
    }
	\caption{The average number of \texttt{cache-misses} during the classification of different categories for \textbf{a)} MNIST and \textbf{b)} CIFAR-10 dataset\label{fig:cache_misses}}
\end{figure}

In order to demonstrate the idea of information leakage we consider two publicly available image datasets \textbf{MNIST}~\cite{lecun2010mnist} and \textbf{CIFAR-10}~\cite{krizhevsky2014cifar} in our study. Most images in these datasets have little or no clutter, and the classification objects tend to be centered in each of the images. Our main speculation from these datasets is that the effect of CNN operations on the hardware activities will be different for different categories, as different images belonging to a particular class activates a set of neuron in the CNN, which might not get activated for other images belonging to a different class. The activation and inactivation of these neurons influence the CNN operation affecting CPU cache, branch predictor and other units differently for different categories. In order to support the claim we monitor the average number of \texttt{cache-misses} (The measurement of these hardware activities is discussed in Section~\ref{sec:measure_hpc} and Section~\ref{sec:method} in more details) for both the dataset while classifying each category through a CNN model and present the result in Figure~\ref{fig:cache_misses}\footnote[2]{Without loss of generality, we present the result for four different categories for both the datasets.}. The figure shows that the average number of \texttt{cache-misses} is different for different categories showing a possible venue for information leakage which depends on the input. The observation in Figure~\ref{fig:cache_misses} motivates us to use the information provided by the low-level hardware events in order to develop our evaluation framework. In the next section, we present a brief discussion on how these hardware activities can be measured in a computing environment using HPCs.

\section{Measuring Hardware Activities using Hardware Performance Counters}\label{sec:measure_hpc}
Hardware Performance Counters (HPCs) are a set of special purpose registers, which are built into most of the modern microprocessor's Performance Monitoring Unit (PMU) to dynamically observe the hardware related activities in a computing environment. These registers can be programmed easily to collect the number of occurrences of different micro-architectural events (like cache misses, branch mispredictions, retired instructions, etc.) during the execution of a program in the processor. The advantage of these performance counters is that they can be accessed very fast without affecting or slowing down any software execution. HPCs can be monitored dynamically using the \texttt{perf} tool, available in Linux kernels 2.6.31 and above, which can be invoked with administrative privilege to access these performance counters with very low granularity. The range of HPC events those can be monitored using the perf tool is more than 1000 depending on the processor Instruction Set Architectures. However, in most of the Linux based systems, the perf tool is limited to observing a maximum of 6 to 8 hardware events in parallel because of the restrictions in the number of built-in HPC registers. Moreover, some of the events are not even supported by all the processors. Hence, we have experimented with some of the basic hardware events which are supported accross the processors. The command to monitor a particular HPC event for a specific process is as follows:

\begin{center}
    \texttt{\textbf{perf stat -e <event\_name> -p <process\_id>}}
\end{center}

These HPCs are widely used as a source of side-channel~\cite{uhsadel2008exploiting,bhattacharya2015watches,alam2017tackling} in order to compromise the security of several mathematically elegant cryptographic encryption algorithms. In the next section, we present the methodology to evaluate the privacy of a CNN model in the presence of information leakages in the form of HPCs.

\section{Methodology for Evaluation}\label{sec:method}
\begin{figure}[!t]
	\centering
	\subfigure[]{
	    \label{fig:threat_model:a}
		\includegraphics[width=0.5\textwidth]{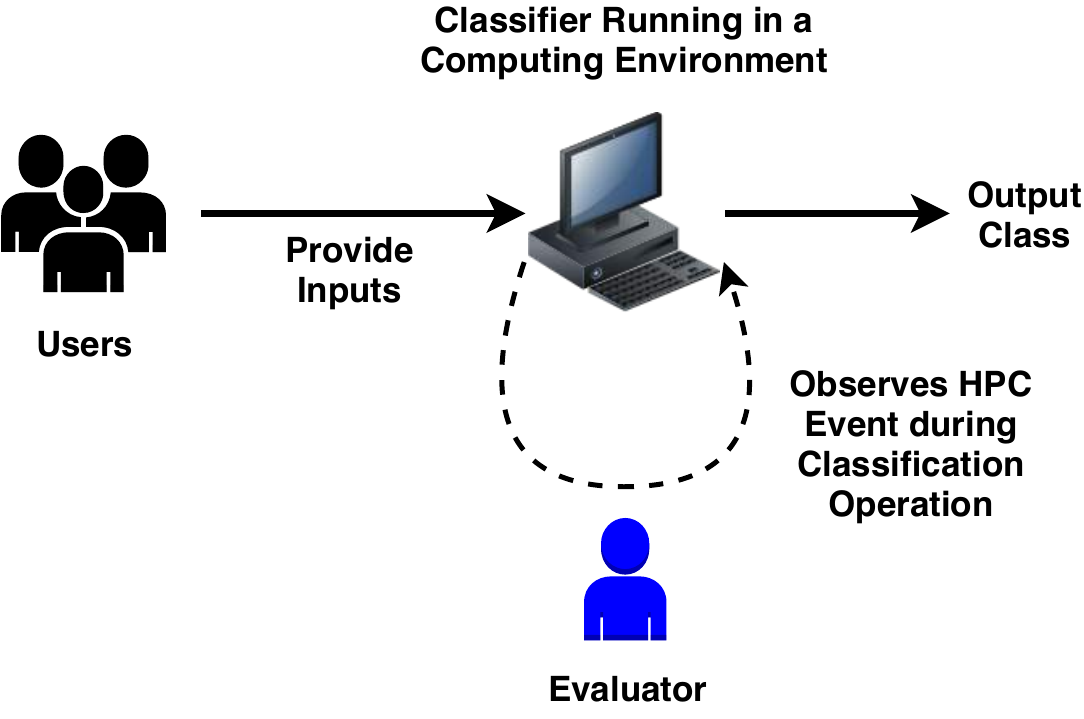}
	}
	\subfigure[]{
	    \label{fig:threat_model:b}
		\includegraphics[width=0.35\textwidth]{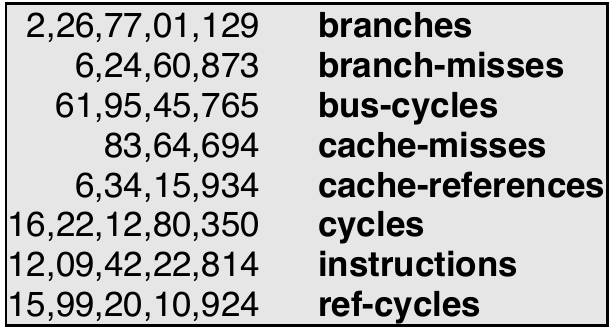}
    }
	\caption{\textbf{Function of the \emph{Evaluator}:} \textbf{a)} The \textit{Evaluator} can only monitor the hardware events of the computing environment during the classification operation for a \textit{User}. \textbf{b)} The values of different hardware events during the classification of a sample MNIST image. The \textit{Evaluator} does not know the input image but can obtain these values.\label{fig:threat_model}}
\end{figure}

Let us consider a scenario where a CNN model, trained on the private information, is executed in a computing environment as shown in Figure~\ref{fig:threat_model:a}. A group of \emph{User} can access this model to get predictions on their respective inputs. Let us also consider the \emph{Evaluator} who is not provided with the details of the CNN model but can monitor the HPCs during the execution of the model by its process id as discussed in Section~\ref{sec:measure_hpc}. The \emph{Evaluator} can only get information as shown in Figure~\ref{fig:threat_model:b}, which indicates the quantitative values for different hardware events in a single classification operation. The evaluator, having the administrative privilege, can be employed to observe the HPC events for different category of images. The operations of the evaluator are as follows:
\begin{enumerate}
    \item It monitors different HPC events in parallel during the classification operation of different categories of input images, considering each category individually. Thus, resulting in distributions of different HPC events for each class of inputs.
    \item It employs a hypothesis testing methodology by computing $t$-statistics on the distributions of same HPC events for different categories.
\end{enumerate}

The evaluator raises the alarm if the \emph{null hypothesis} is rejected, which signifies that the distributions are different. If the distributions of an HPC event for different inputs are not distinguishable from each other, we say that an adversary will not be able to exploit this side-channel information in order to uncover the private input images, thus indicating an efficient implementation of the CNN model.

In the next section, we evaluate the security of a CNN model considering MNIST and CIFAR-10 dataset as mentioned previously.

\section{Results}
\subsection{Experimental Setup}
We have implemented two CNN models for MNIST and CIFAR-10 dataset using the \texttt{tensorflow} library for our evaluation. We executed the models in Intel Xeon E5-2690 CPU having Ubuntu 18.04 with 4.15.0-36-generic kernel. All the hardware related activities are measured using the \texttt{perf} tool as mentioned previously.

\subsection{Case Study on MNIST}
\begin{figure}[!t]
	\begin{adjustwidth}{-1cm}{-1cm}
	\centering
	\subfigure[]{
	    \label{fig:distribution:a}
		\includegraphics[width=0.5\textwidth]{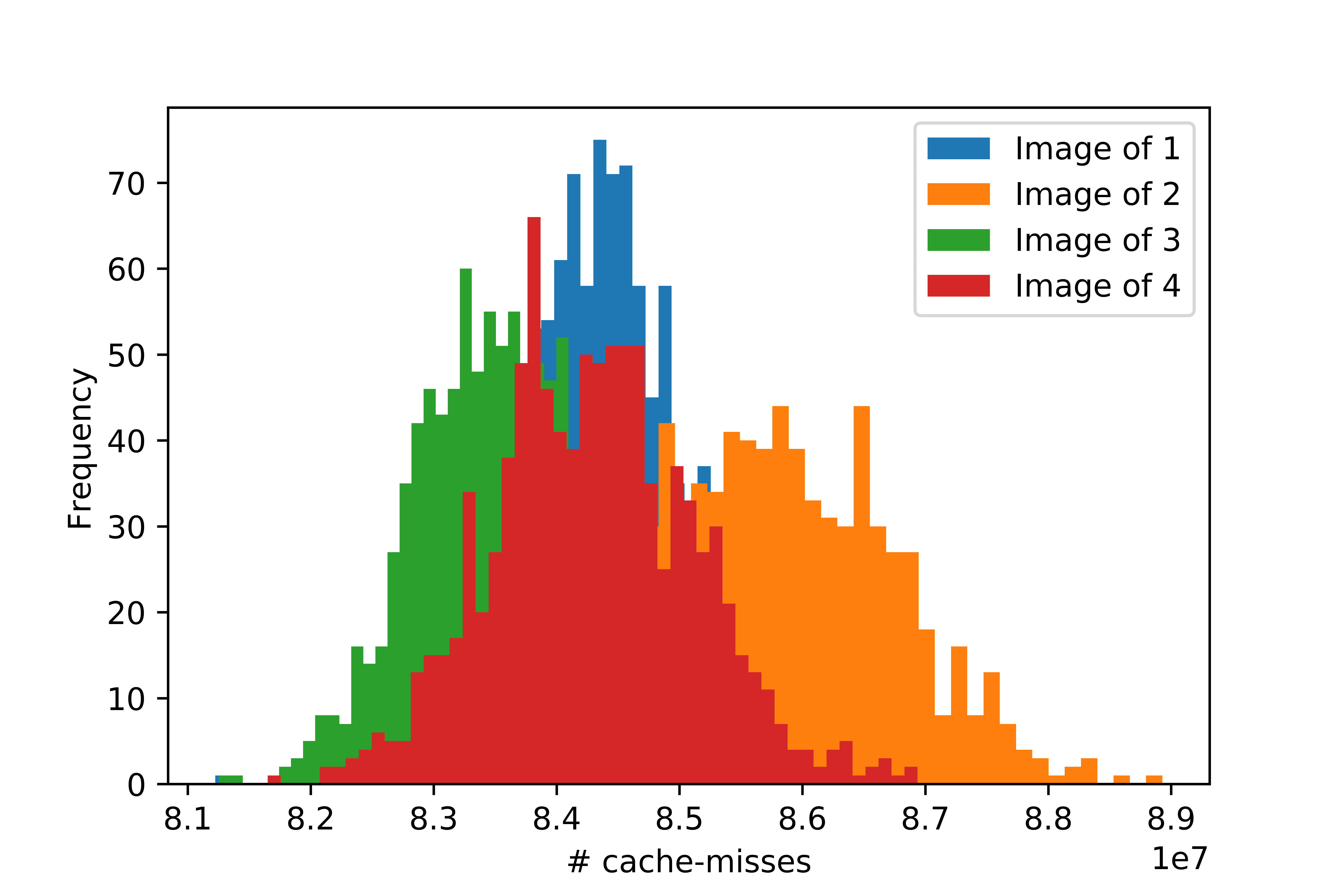}
	}
	\subfigure[]{
	    \label{fig:distribution:b}
		\includegraphics[width=0.5\textwidth]{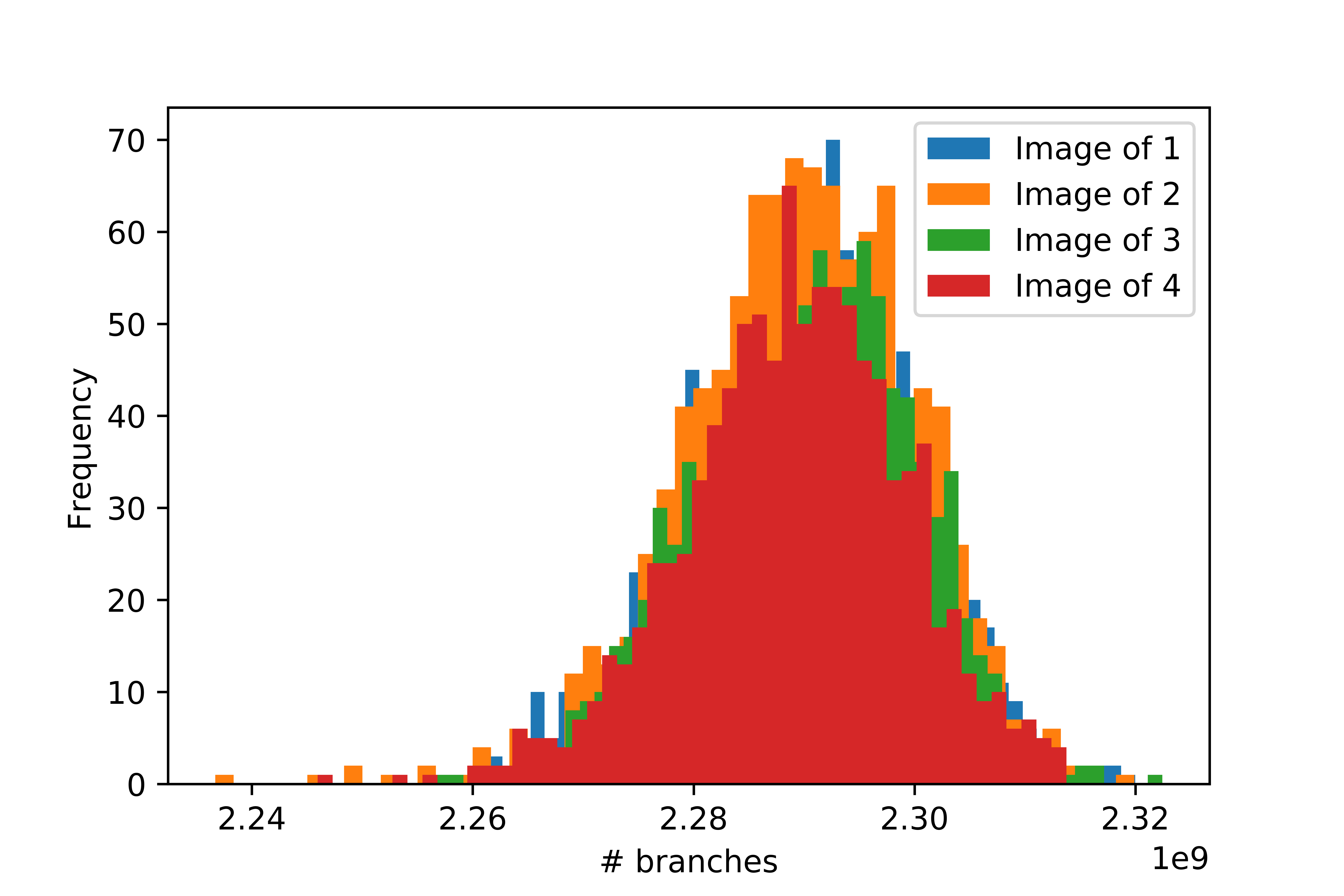}
    }
	\caption{Distributions of HPC events \textbf{a)} \texttt{cache-misses} and \textbf{b)} \texttt{branches} during the testing operation for different categories of images present in MNIST dataset.\label{fig:distribution}}
	\end{adjustwidth}
\end{figure}

We monitored different hardware events during the prediction of each categories for MNIST dataset. We observed that some of the events can produce different distributions for different categories. The distributions of events \texttt{cache-misses} and \texttt{branches} for all the test images belonging to different categories are shown in Figure~\ref{fig:distribution}. If we apply $t$-test on the distributions shown in Figure~\ref{fig:distribution:a}, we can easily distinguish them. The $t$ and $p$ values for the tests are shown in Table~\ref{tab:mnist}. The $t$-test is conducted with $95\%$ confidence interval. $t_{i, j}$ signifies the $t$-test on the distribution for category $i$ and category $j$. The \textbf{bold} faced results indicates that the two categories can be distinguished using the $t$-test. All the distributions shown in Figure~\ref{fig:distribution:b} can not be distinguished using the $t$-test, but some of the categories can be. The results are presented in Table~\ref{tab:mnist}, which shows that the $p$-values are significantly higher for most of the results. Hence, the event \texttt{cache-misses} leaks valuable information on all the input categories and the event \texttt{branches} can be exploited to distinguish the inputs belonging to category 2 and category 3, which triggers the evaluator to raise an alarm.

\begin{figure}[!t]
    \begin{adjustwidth}{-1cm}{-1cm}
	\centering
	\subfigure[]{
	    \label{fig:caltech_distribution:a}
		\includegraphics[width=0.5\textwidth]{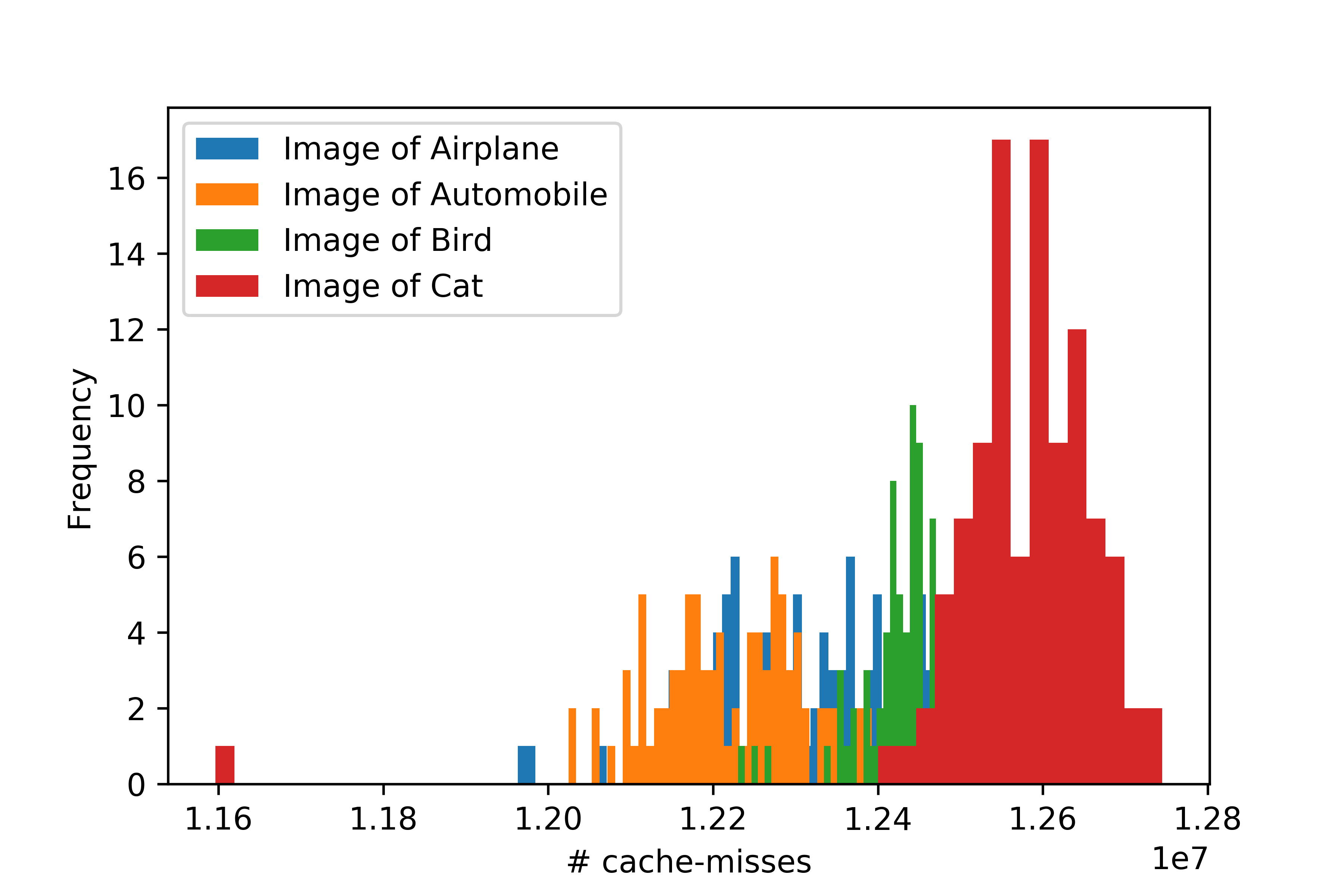}
	}
	\subfigure[]{
	    \label{fig:caltech_distribution:b}
		\includegraphics[width=0.5\textwidth]{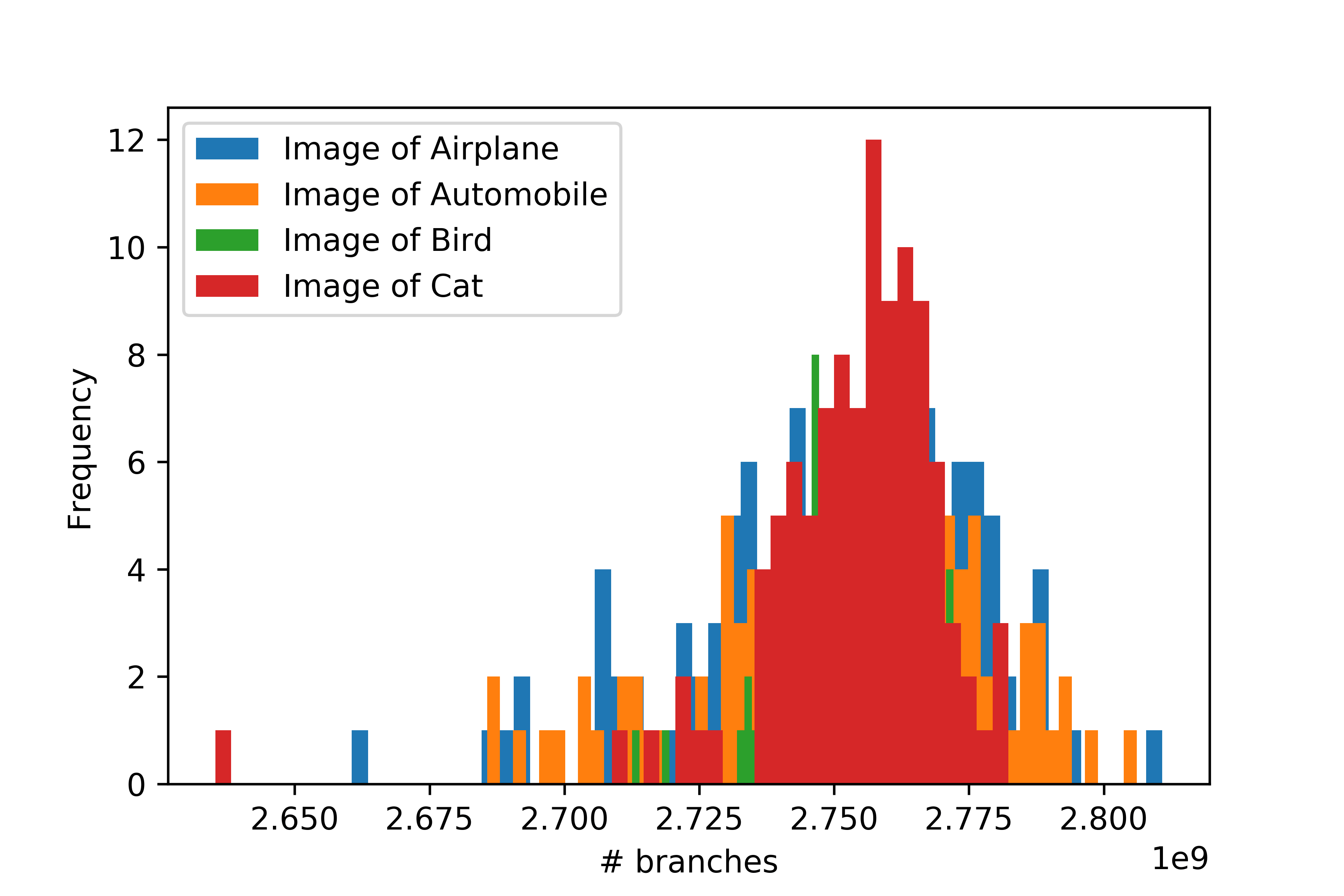}
    }
	\caption{Distributions of HPC events \textbf{a)} \texttt{cache-misses} and \textbf{b)} \texttt{branches} during the testing operation for different categories of images present in CIFAR-10 dataset.\label{fig:caltech_distribution}}
	\end{adjustwidth}
\end{figure}

\begin{table}[!t]
\begin{minipage}[b]{0.5\textwidth}\centering
\caption{Results of $t$-test on the distributions obtained from the HPC events \texttt{cache-misses} and \texttt{branches} for MNIST dataset\label{tab:mnist}}
\adjustbox{max width=0.95\textwidth}{%
\begin{tabular}{c|c|c|c|c|}
\cline{2-5}
\textbf{}                 & \multicolumn{2}{c|}{\texttt{cache-misses}} & \multicolumn{2}{c|}{\texttt{branches}} \\ \hline
\multicolumn{1}{|c|}{}    & $t$-values              & $p$-values           & $t$-values           & $p$-values          \\ \hline
\multicolumn{1}{|c|}{$t_{1,2}$} & \textbf{-21.8166}     & \textbf{$\approx$0}         & 0.4303             & 0.6669            \\ \hline
\multicolumn{1}{|c|}{$t_{1,3}$} & \textbf{-25.7566}     & \textbf{$\approx$0}         & 1.6565             & 0.0977            \\ \hline
\multicolumn{1}{|c|}{$t_{1,4}$} & \textbf{2.5334}       & \textbf{0.0113}    & 0.9537             & 0.3403            \\ \hline
\multicolumn{1}{|c|}{$t_{2,3}$} & \textbf{40.5268}      & \textbf{$\approx$0}         & \textbf{-2.0064}   & \textbf{0.0449}   \\ \hline
\multicolumn{1}{|c|}{$t_{2,4}$} & \textbf{22.6505}      & \textbf{$\approx$0}         & 0.4941             & 0.6212            \\ \hline
\multicolumn{1}{|c|}{$t_{3,4}$} & \textbf{-20.9758}     & \textbf{$\approx$0}         & \textbf{2.5435}    & \textbf{0.0110}   \\ \hline
\end{tabular}}
\end{minipage}
\hspace{0.1cm}
\begin{minipage}[b]{0.5\textwidth}\centering
\caption{Results of $t$-test on the distributions obtained from the HPC events \texttt{cache-misses} and \texttt{branches} for CIFAR-10 dataset\label{tab:caltech}}
\adjustbox{max width=0.95\textwidth}{%
\begin{tabular}{c|c|c|c|c|}
\cline{2-5}
\textbf{}                 & \multicolumn{2}{c|}{\texttt{cache-misses}} & \multicolumn{2}{c|}{\texttt{branches}} \\ \hline
\multicolumn{1}{|c|}{}    & $t$-values              & $p$-values           & $t$-values           & $p$-values          \\ \hline
\multicolumn{1}{|c|}{$t_{1,2}$} & \textbf{4.4643}  & \textbf{0.0001}  & -0.8796 & 0.3801            \\ \hline
\multicolumn{1}{|c|}{$t_{1,3}$} & \textbf{11.0415}  & \textbf{$\approx$0}  & \textbf{2.0810} & \textbf{0.0392}            \\ \hline
\multicolumn{1}{|c|}{$t_{1,4}$} & \textbf{-16.3093}  & \textbf{$\approx$0}  & -1.7474 & 0.0823            \\ \hline
\multicolumn{1}{|c|}{$t_{2,3}$} & \textbf{-16.9589}  & \textbf{$\approx$0}  & -1.0332 & 0.3032   \\ \hline
\multicolumn{1}{|c|}{$t_{2,4}$} & \textbf{-21.2428}  & \textbf{$\approx$0}  & -0.7535 & 0.4521            \\ \hline
\multicolumn{1}{|c|}{$t_{3,4}$} & \textbf{-8.4637}  & \textbf{$\approx$0}  & 0.2997 & 0.7647   \\ \hline
\end{tabular}}
\end{minipage}
\end{table}

\subsection{Case Study on CIFAR-10}
We also perform the same experimentation for CIFAR-10 dataset. The distributions of events \texttt{cache-misses} and \texttt{branches}, in this case, are shown in Figure~\ref{fig:caltech_distribution}. The results of $t$-tests for the distributions shown in Figure~\ref{fig:caltech_distribution:a} and Figure~\ref{fig:caltech_distribution:b} are presented in Table~\ref{tab:caltech}. In this case also we can distinguish between all the distributions if we consider the event \texttt{cache-misses} and the distributions of category 1 and category 3 can be distinguished using the evant \texttt{branches}. This triggers the evaluator to raise an alarm about the information leakage related to the input image.

\section{Conclusion and Future Work}
In this work, we presented a strategy to evaluate the data privacy of deep neural network architectures with readily available tools. We took the aid of low-level HPC events and $t$-test in designing the evaluation strategy. We presented the result based on a CNN based image classifier on two publicly available datasets, MNIST and CIFAR-10. Our evaluation tool highlights the need for designing CNN architectures with indistinguishable CPU footprints while classifying different image categories in order to implement a privacy preserving classifier. In a future scope of this work, we would also like to explore the vulnerabilities in other deep learning models with different application scenarios.

\bibliographystyle{unsrt}
\bibliography{ref}

\end{document}